\documentclass{article}

\usepackage[final]{tackling_climate_workshop_style}

\usepackage[utf8]{inputenc} 
\usepackage[T1]{fontenc}    
\usepackage{hyperref}       
\usepackage{url}            
\usepackage{booktabs}       
\usepackage{amsfonts}       
\usepackage{amsmath}
\usepackage{nicefrac}       
\usepackage{microtype}      
\usepackage{bm,mathtools}

\title{Learning to forecast diagnostic parameters using pre-trained weather embedding}

\author{%
  Peetak Mitra \\
  Excarta\\
  \texttt{peetak@excarta.io} \\
   \And
   Vivek Ramavajjala \\
   Excarta \\
   \texttt{vivek@excarta.io} \\
}

\begin{document}

\maketitle

\begin{abstract}

Data-driven weather prediction (DDWP) models are increasingly becoming popular for weather forecasting. However, while operational weather forecasts predict a wide variety of weather variables, DDWPs currently forecast a specific set of key prognostic variables. Non-prognostic ("diagnostic") variables are sometimes modeled separately as dependent variables of the prognostic variables (c.f. FourCastNet \cite{pathak2022fourcastnet}), or by including the diagnostic variable as a target in the DDWP. However, the cost of training and deploying bespoke models for each diagnostic variable can increase dramatically with more diagnostic variables, and limit the operational use of such models. Likewise, retraining an entire DDWP each time a new diagnostic variable is added is also cost-prohibitive. We present an two-stage approach that allows new diagnostic variables to be added to an end-to-end DDWP model without the expensive retraining. In the first stage, we train an autoencoder that learns to embed prognostic variables into a latent space. In the second stage, the autoencoder is frozen and "downstream" models are trained to predict diagnostic variables using only the latent representations of prognostic variables as input. Our experiments indicate that models trained using the two-stage approach offer accuracy comparable to training bespoke models, while leading to significant reduction in resource utilization during training and inference. This approach allows for new "downstream" models to be developed as needed, without affecting existing models and thus reducing the friction in operationalizing new models.

\end{abstract}

\section{Introduction}

\begin{figure}
    \centering
    \includegraphics[width=0.9\linewidth]{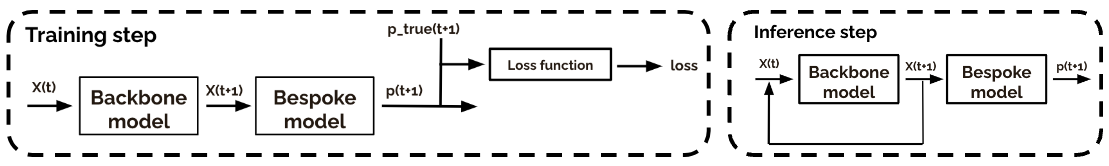}
    \caption{FourCastNet used a bespoke model to predict precipitation. A backbone model that only predicts "core" weather state, and a bespoke model that uses the weather state to predict diagnostic variables. Scaling such an approach for many downstream tasks would be challenging.}
    \label{fig:bespoke}
\end{figure}

In recent years, data-driven weather prediction (DDWP) models, such as FourCastNet \cite{pathak2022fourcastnet}, GraphCast \cite{lam2022graphcast}, and PanguWeather \cite{bi2023accurate}, have shown remarkable skill in multi-day forecasts compared to state-of-the-art, operational Numerical Weather Prediction (NWP) models such as Integrated Forecast Service (IFS) \cite{rasp2023weatherbench}, and reanalyses data such as ERA5 \cite{nguyen2023climax}. Recent studies have indicated that these DDWP models and NWP emulators, while fully data-dependent, have learnt useful physics proving the robustness of this approach \cite{hakim2023dynamical}, and behave similarly to NWPs when comparing against \textit{in-situ} observations \cite{ramavajjala2023verification}. However, most DDWPs often do not focus on diagnostic meteorological parameters such as precipitation, total cloud cover, solar irradiance, soil moisture, etc., that are essential for decision making in industries like transportation, energy, and agriculture. GraphCast\cite{lam2022graphcast} forecasts precipitation along with other prognostic variables, while FourCastNet\cite{pathak2022fourcastnet} trains a bespoke model to predict precipitation, with the same complexity as the "backbone" model itself (Figure \ref{fig:bespoke}). In either approach, a full-scale DDWP needs to be re-trained for each new diagnostic variable to be predicted, which is not scalable either in training or in inference.

It is commonly observed that the first few layers of most deep learning models learn a dense latent representation of the high-dimensional input space\cite{li2023transformers, zeiler2014visualizing, zhou2022understanding}. Since in most cases, the starting state for predict diagnostic variables is the current set of prognostic variables, we propose a two-stage approach to modeling diagnostic variables: first, we train an auto-encoder that learns to embed prognostic variables in a dense latent space. Then, for each diagnostic variable we train a "downstream" model that uses the dense representation of prognostic variables as an input. As the auto-encoder has already learned a high-quality dense representation, we can significantly reduce the size of each "downstream" model. The closest such approach is proposed in W-MAE \cite{man2023w}, which also trains an auto-encoder but fine-tunes the auto-encoder to forecast prognostic variables and precipitation. Fine tuning the encoder prevents new diagnostic variables from being modeled without also re-training all previous models.

In contrast, we propose training an autoencoder using decades of high quality ground truth hourly reanalyses data from ERA5, which is then frozen, and using the frozen auto-encoder to train downstream tasks. The encoder's dense representations can be used as-is, or be combined with other task-specific signals to directly forecast industry-specific weather-dependent outcomes, e.g., predicting likelihood of pest outbreaks as a function of humid weather conditions.

\subsection{Our contributions}

The main contributions of this study are in demonstrating that (i) complex, high-dimensional weather data can be effectively represented in a condensed, embedded representation at a moderate cost, and (ii) weather-dependent tasks can be learned efficiently from dense representations of weather. We demonstrate this by not only quantifying the reconstruction error of the autoencoder model, but using it to build task-specific downstream applications, and comparing their performance to bespoke models. This finding unlocks the ability to readily use the weather state in key decision making processes by dramatically reducing the cost of training and deploying ML-driven weather models.

\begin{figure}[ht]
    \centering
    \includegraphics[width=0.6\linewidth]{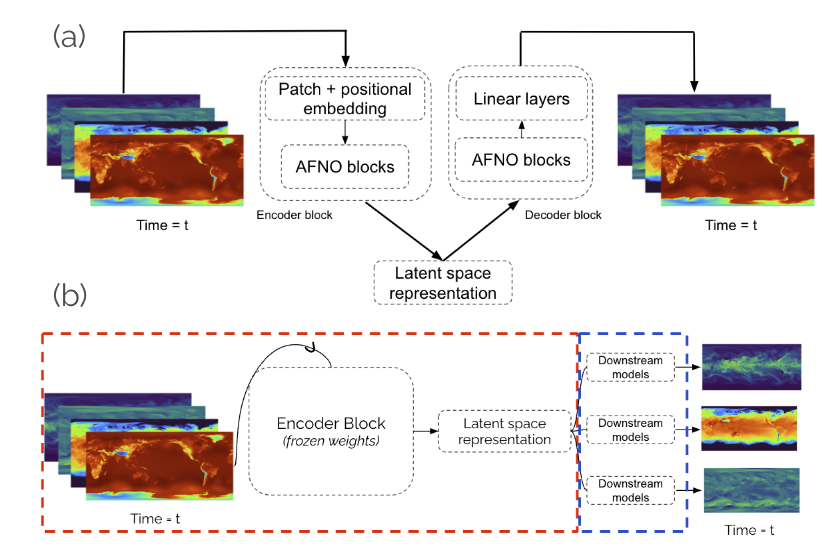}
    \caption{\textbf{a}: The autoencoder model learns, via reconstructing the input, a rich dense representation of the current weather state, and \textbf{b}: The trained encoder is frozen and used to produce latent representations of the current weather state (block in \textit{red}), which are used as inputs to train smaller downstream branches (block in \textit{blue}).}
    \label{fig:schematic-label}
\end{figure}

\section{Experimental design}
Our objective is to demonstrate that the two-stage modeling approach described above provides performance comparable to training bespoke, full-scale DDWPs for diagnostic variables.

\subsection{Dataset}

The dataset used for this study is the publicly available reanalyses data product, ERA5 \cite{hersbach2020era5}. We choose a subset of the variables (defined in Table \ref{var-table}) provided in ERA5 which are commonly considered to be prognostic variables, and thus critical to capturing the essential information about the current weather state. These variables are then normalized to zero mean and unit variance using climatological statistics computed over decades of data. In our experiments, we focus on modeling total cloud cover ($\mathrm{tcc}$), and top-level soil temperature ($\mathrm{stl1}$), which are critical for operational decisions in the energy and agriculture industry, respectively. While our approach was demonstrated on these variables for this study, this approach is task-agnostic and can readily be extended to model any weather or non-weather variables.

\subsubsection{Input and target variables}

The 54 prognostic input variables, used at 0.25\textdegree resolution, are similar to the parameters used in other DDWP studies \cite{pathak2022fourcastnet, bi2023accurate}.  

\begin{table}[ht]
  \caption{Model inputs, totalling 54 weather variables}
  \label{var-table}
  \centering
  \begin{tabular}{ll}
    \toprule
    Vertical Level    & Variables   \\
    \midrule
    Surface & $\mathrm{u_{10}, v_{10}, t_{2m}, d_{2m}, msl, sp, u_{100}, v_{100}, tcwv}$  \\
    \midrule
   1000, 925, 850, 700, 500, 300, 250, 200, 50  (all in hPa)  & $\mathrm{t, u,  v, z, r}$     \\
    \bottomrule
  \end{tabular}
\end{table}

The target variables are listed in Table \ref{outvar-table}. While the modeling approach is variable-agnostic, we choose these two variables because of their relative importance across the energy and agriculture industries, and also their relevant modeling challenges. For example, total cloud cover ($\textit{tcc}$) is influenced by conditions at all levels of the atmosphere, while top-level soil temperature ($\textit{stl1}$) is influenced more heavily by surface weather. For evaluation, we use the RMSE metric for both total cloud cover and soil temperature. For total cloud cover, we additionally use the structural similarity metric (SSIM) to measure the spatial similarity in the predicted and true cloud cover.

\begin{table}[ht]
  \caption{Target variables used in the study }
  \label{outvar-table}
  \centering
  \begin{tabular}{llll}
    \toprule
    Variable & Data Range &Activation & Masks  \\
    \midrule
    $\mathrm{stl1}$  & 220-290 K &  None & land-sea \\
    \midrule
    $\mathrm{tcc}$  & 0-1 &  softmax & None \\
    \bottomrule
  \end{tabular}
\end{table}

\subsection{Model architecture}
Our two-stage training approach is as follows:
\begin{enumerate}
    \item Train an auto-encoder on ERA5 data to learn a dense representation of the prognostic variables at time \textit{t}
    \item Train task-specific models to predict diagnostic variables using the dense representation as input.
\end{enumerate}
Since only the encoder model sees the prognostic variables, and the encoder does not make assumptions about the prognostic variables, the above setup can be coupled with any "backbone" forecasting models as long as the necessary prognostic variables are available. In other words, this approach can be coupled with GraphCast, Pangu-Weather, or any DDWP, to automatically predict additional diagnostic variables.

\subsection{Autoencoder}

The weather state embedding is built using an autoencoder which consists of two different modules, an encoder ($\phi$) that represents the input data [$\mathrm{B, C_{in}, H_{in}, W_{in}}$] into a dense, latent representation [$\mathrm{B, L, H_{L}, W_{L}}$], functionally shown as
\begin{equation}
    \phi : \chi_{t} \xrightarrow{}  F 
\end{equation}

and a decoder ($\psi$) which uses the latent representation, to recreate the original data [$\mathrm{B, C_{out}, H_{out}, W_{out}}$], functionally represented as
\begin{equation}
    \psi: F \xrightarrow{} \chi_{t}
\end{equation}

During training, the overall minimization process updates the encoder and decoder weights simultaneously, until fully trained.
\begin{equation}
    \phi, \psi = \texttt{arg min}_{\phi, \psi} \lVert \chi_{t} - (\phi \circ \psi) \chi_{t} \rVert ^2
\end{equation}

Once trained, only the encoder is used to embed the current weather state into a dense representation. In our study, we use construct the autoencoder using AFNO layers as described in FourCastNet \cite{pathak2022fourcastnet}. However, this approach is architecture-agnostic, meaning other architectures like Vision Transformers (ViT) or graph-based methods can similarly be used. Table \ref{model-table} defines the architecture choices and total model sizes.

\subsection{Task-specific downstream model}

The task specific downstream models ($\mathrm{M}$) use the dense representations from the auto-encoder as the input, functionally seen as $\mathrm{Y_{t} = M(\phi(\chi_{t}))}$.

\subsection{Bespoke models}

To enable a fair comparison of the performance of the task-specific downstream model, we train bespoke models based on the FourCastNet backbone model architecture \cite{pathak2022fourcastnet}, using the same prognostic variables as input (c.f. Table \ref{var-table}) and the same diagnostic variables (c.f. Table \ref{outvar-table}) as output. All models use standard Adam \cite{kingma2014adam} optimizer with an exponential learning rate decay of an initial rate of 0.0002.

The following table shows the model sizes used in the study, and we note that the "downstream" models are significantly smaller than the bespoke models.

\begin{table}[ht]
  \caption{Model architecture choices and total number of parameters}
  \label{model-table}
  \centering
  \begin{tabular}{lllll}
    \toprule
    \multicolumn{5}{c}{AFNO Hyperparameters}                   \\
    \cmidrule(r){2-5}
    Model     & \# of layers &  dim  & patch size   & Total \# of parameters \\
    \midrule
    Autoencoder        & 4 each for encoder, decoder   & 768  & 8  & 49M     \\
    Downstream models  & 6        & 768  & 8 & 28M     \\
    Bespoke models     & 12       & 768  & 8 & 75M  \\
    \bottomrule
  \end{tabular}
\end{table}

\section{Evaluations and Discussions}

 The evaluations were performed over the hold-out test year of 2020. To holistically characterize the model performance and account for seasonality, we evaluate the performance on the 1st, 2nd, 15th and 16th days of each month. For each date, evaluations are performed for each hour of the day to account for the time-of-day effect. The forecasts are evaluated by comparing the predictions against ERA5 data using two metrics: the root mean squared error (RMSE) and the structural similarity index measure (SSIM), computed hourly over the entire latitude-longitude grid. The mean and standard deviation are computed by collating the data from across all evaluation periods. While the RMSE provides a global average of error, the SSIM metric is a qualitative, reconstructive, perceptual metric that quantifies image quality degradation to the ground truth data.

\subsection{Autoencoder}

Autoencoders often produce manifolds that over fit to noisy training data  \cite{lee2021neighborhood}. Such over fitting could lead to a poor starting state for the task-specific fine tuning models degrading their performance. To prevent overfitting and understand the impact of model architecture choices, we performed ablation studies on the patch size, number of AFNO layers, and the number of channels used in the autoencoder. Results indicate that reducing the patch size leads to slight improvement in performance, but the training and inference cost scales quadratically for when doubling patch sizes, leading to a significantly unfavorable compute-accuracy trade off for smaller patch sizes. For this study, we use the encoder with a patch size of 8, given its lower computational overhead and overall similar performance to the patch size 4 models. The number of AFNO layers and channels are by far the most important hyperparameters (c.f. Table \ref{autoencresults-table}), in line with the expectations that an over-parameterized model can more easily overfit.

\begin{table}[ht]
  \caption{Ablation study for autoencoder model} 
  \label{autoencresults-table}
  \centering
  \begin{tabular}{llll}
    \toprule
    Patch size, \# Layers    & variable & Avg. RMSE: $\mu$, $\sigma$    &  Avg. SSIM: $\mu$, $\sigma$  \\
    \midrule
    patch:4, layers:8   & u10  & 0.187, 0.0023  & 0.9947, 0.0004   \\
    patch:4, layers:4   & u10 & \textbf{0.128,  0.0015}  & \textbf{0.9973, 0.0002}     \\
    patch:8, layers:8   & u10 & 0.185, 0.0010  & 0.9937, 0.0005     \\
    patch:8, layers:4   & u10 & 0.134, 0.0016  & 0.9947, 0.0004    \\
    \midrule
    patch:4, layers:8   & t2m  & 0.676, 0.0344  & 0.9920, 0.0005   \\
    patch:4, layers:4   & t2m & 0.481, 0.0221  & \textbf{0.9944, 0.0004}    \\
    patch:8, layers:8   & t2m & 0.581, 0.0114  & 0.9877, 0.0007    \\
    patch:8, layers:4   & t2m & \textbf{0.454, 0.0105}  & 0.9886, 0.0007   \\
    \bottomrule
  \end{tabular}
\end{table}

\subsection{Inter model comparisons}

The starting point for the task-specific downstream model is the dense representation from the trained encoder model. In the table \ref{results-table}, we compare the average RMSE and SSIM scores for the two modeled variables between the two separate modeling approaches. Results indicate that the average errors between the bespoke and "downstream" models are almost nearly identical, further validating the two-stage modeling approach. However, each downstream model is roughly half the size of the corresponding bespoke models, and during inference the auto-encoder needs to be run only once for any number of downstream tasks. The significantly lower overhead during training, and dramatically lower footprint during inference, makes the cost-accuracy trade off hugely favorable for the smaller downstream models.

\begin{table}[ht]
  \caption{Comparing model performance on diagnostic variables} 
  \label{results-table}
  \centering
  \begin{tabular}{llll}
    \toprule
    Model type    & variable & Avg. RMSE: $\mu$, $\sigma$    &  Avg. SSIM: $\mu$, $\sigma$  \\
    \midrule
    Bespoke   & tcc  & \textbf{0.1656, 0.0076}  & 0.5648, 0.0134  \\
    Downstream   & tcc & 0.1677, 0.0084  & \textbf{0.5926, 0.0141}   \\
    \midrule
    Bespoke   & stl1 & \textbf{11.761, 0.14367}  & \textbf{0.9633, 0.00034}  \\
    Downstream   & stl1 & 11.784, 0.1645  & 0.9632, 0.00036\\
    \bottomrule
  \end{tabular}
\end{table}



\section{Conclusions}

While DDWPs have steadily improved their forecast skill, their use in operational forecasts is limited if only prognostic variables are predicted. Many decisions in the real world are dependent on diagnostic variables, and neither retraining DDWPs nor building bespoke models for each new diagnostic variables are scalable. The main impetus of this work is to develop efficient, scalable ways to train and operationalize new models dependent on weather. We show that prognostic variables can be embedded in a task-agnostic latent space, and be used as a starting state to train task-specific downstream models without any appreciable drop in model accuracy, and significant savings in computational costs.

\newpage

\section{Data and code availability}

The authors are working to release an extended paper with more evaluations and open source the trained model weights. In the meantime, if you have any questions please reach out to us at contact@excarta.io.

\bibliographystyle{apalike}
\bibliography{refs}

\begin{thebibliography}{}

\bibitem[Bi et~al., 2023]{bi2023accurate}
Bi, K., Xie, L., Zhang, H., Chen, X., Gu, X., and Tian, Q. (2023).
\newblock Accurate medium-range global weather forecasting with 3d neural networks.
\newblock {\em Nature}, pages 1--6.

\bibitem[Hakim and Masanam, 2023]{hakim2023dynamical}
Hakim, G.~J. and Masanam, S. (2023).
\newblock Dynamical tests of a deep-learning weather prediction model.
\newblock {\em arXiv preprint arXiv:2309.10867}.

\bibitem[Hersbach et~al., 2020]{hersbach2020era5}
Hersbach, H., Bell, B., Berrisford, P., Hirahara, S., Hor{\'a}nyi, A., Mu{\~n}oz-Sabater, J., Nicolas, J., Peubey, C., Radu, R., Schepers, D., et~al. (2020).
\newblock The {ERA5} global reanalysis.
\newblock {\em Quarterly Journal of the Royal Meteorological Society}, 146(730):1999--2049.

\bibitem[Kingma and Ba, 2014]{kingma2014adam}
Kingma, D.~P. and Ba, J. (2014).
\newblock Adam: A method for stochastic optimization.
\newblock {\em arXiv preprint arXiv:1412.6980}.

\bibitem[Lam et~al., 2022]{lam2022graphcast}
Lam, R., Sanchez-Gonzalez, A., Willson, M., Wirnsberger, P., Fortunato, M., Pritzel, A., Ravuri, S., Ewalds, T., Alet, F., Eaton-Rosen, Z., Hu, W., Merose, A., Hoyer, S., Holland, G., Stott, J., Vinyals, O., Mohamed, S., and Battaglia, P. (2022).
\newblock Graphcast: Learning skillful medium-range global weather forecasting.

\bibitem[Lee et~al., 2021]{lee2021neighborhood}
Lee, Y., Kwon, H., and Park, F. (2021).
\newblock Neighborhood reconstructing autoencoders.
\newblock {\em Advances in Neural Information Processing Systems}, 34:536--546.

\bibitem[Li et~al., 2023]{li2023transformers}
Li, Y., Li, Y., and Risteski, A. (2023).
\newblock How do transformers learn topic structure: Towards a mechanistic understanding.
\newblock {\em arXiv preprint arXiv:2303.04245}.

\bibitem[Man et~al., 2023]{man2023w}
Man, X., Zhang, C., Li, C., and Shao, J. (2023).
\newblock W-mae: Pre-trained weather model with masked autoencoder for multi-variable weather forecasting.
\newblock {\em arXiv preprint arXiv:2304.08754}.

\bibitem[Nguyen et~al., 2023]{nguyen2023climax}
Nguyen, T., Brandstetter, J., Kapoor, A., Gupta, J.~K., and Grover, A. (2023).
\newblock Climax: A foundation model for weather and climate.
\newblock {\em arXiv preprint arXiv:2301.10343}.

\bibitem[Pathak et~al., 2022]{pathak2022fourcastnet}
Pathak, J., Subramanian, S., Harrington, P., Raja, S., Chattopadhyay, A., Mardani, M., Kurth, T., Hall, D., Li, Z., Azizzadenesheli, K., Hassanzadeh, P., Kashinath, K., and Anandkumar, A. (2022).
\newblock Four{C}ast{N}et: A {G}lobal {D}ata-driven {H}igh-resolution {W}eather {M}odel using {A}daptive {F}ourier {N}eural {O}perators.

\bibitem[Ramavajjala and Mitra, 2023]{ramavajjala2023verification}
Ramavajjala, V. and Mitra, P.~P. (2023).
\newblock Verification against in-situ observations for data-driven weather prediction.
\newblock {\em arXiv preprint arXiv:2305.00048}.

\bibitem[Rasp et~al., 2023]{rasp2023weatherbench}
Rasp, S., Hoyer, S., Merose, A., Langmore, I., Battaglia, P., Russel, T., Sanchez-Gonzalez, A., Yang, V., Carver, R., Agrawal, S., et~al. (2023).
\newblock Weatherbench 2: A benchmark for the next generation of data-driven global weather models.
\newblock {\em arXiv preprint arXiv:2308.15560}.

\bibitem[Zeiler and Fergus, 2014]{zeiler2014visualizing}
Zeiler, M.~D. and Fergus, R. (2014).
\newblock Visualizing and understanding convolutional networks.
\newblock In {\em Computer Vision--ECCV 2014: 13th European Conference, Zurich, Switzerland, September 6-12, 2014, Proceedings, Part I 13}, pages 818--833. Springer.

\bibitem[Zhou et~al., 2022]{zhou2022understanding}
Zhou, D., Yu, Z., Xie, E., Xiao, C., Anandkumar, A., Feng, J., and Alvarez, J.~M. (2022).
\newblock Understanding the robustness in vision transformers.
\newblock In {\em International Conference on Machine Learning}, pages 27378--27394. PMLR.

\end{thebibliography}
\end{document}